\documentclass{svjour3}
\usepackage{amssymb}
\usepackage{mathtools}
\usepackage[utf8]{inputenc}
\usepackage{multirow}
\usepackage{url}
\usepackage{moresize}
\newcommand{\citet}{\cite}
\usepackage{graphicx}
\usepackage{makecell}
\newcommand{\thinhline}{\Xhline{0.1\arrayrulewidth}}
\usepackage[framemethod=TikZ]{mdframed}
\global\mdfdefinestyle{mdstyle}{linecolor=gray,middlelinewidth=0.5pt,leftmargin=0.0cm,rightmargin=0.0cm,roundcorner=15}
\usepackage{hyperref}

\usepackage{algorithm}
\usepackage[noend]{algpseudocode}
\usepackage{setspace}
\let\Algorithm\algorithm
\renewcommand\algorithm[1][]{\Algorithm[#1]\setstretch{1.1}}

\algrenewcommand{\algorithmiccomment}[1]{\hskip3px$\#$ #1}
\algdef{SE}[SUBALG]{Indent}{EndIndent}{}{\algorithmicend\ }%
\algtext*{Indent}
\algtext*{EndIndent}

\journalname{SN Computer Science}

\usepackage{moresize}

\renewenvironment{quote}{%
  \list{}{%
    \leftmargin0.5cm   
    \rightmargin\leftmargin
  }
  \item\relax
}
{\endlist}

\begin{document}

\title{Neural Networks with \`A La Carte Selection of Activation Functions}

\author{Moshe Sipper}

\institute{M. Sipper \at
              Department of Computer Science, Ben-Gurion University, Beer Sheva 84105, Israel \\
              \email{sipper@gmail.com}, \url{http://www.moshesipper.com/}\\
              \copyright The Author(s), under exclusive licence to Springer Nature Singapore Pte Ltd 2021}


\date{Received: \today / Accepted: date}

\maketitle

\begin{abstract}
Activation functions (AFs), which are pivotal to the success (or failure) of a neural network, have received increased attention in recent years, with researchers seeking to design novel AFs that improve some aspect of network performance. 
In this paper we take another direction, wherein we combine a slew of known AFs into successful architectures, proposing three methods to do so beneficially: 1) generate AF architectures at random, 2) use Optuna, an automatic hyper-parameter optimization software framework, with a Tree-structured Parzen Estimator (TPE) sampler, and 3) use Optuna with a Covariance Matrix Adaptation Evolution Strategy (CMA-ES) sampler. We show that all methods often produce significantly better results for 25 classification problems when compared with a standard network composed of ReLU hidden units and a softmax output unit. 
Optuna with the TPE sampler emerged as the best AF architecture-producing method.

\keywords{Artificial neural network \and Activation function}
\end{abstract}





\begin{quote}
\small
\textit{\`a la carte}: according to a menu or list that prices items separately\\
cf. \textit{table d'h\^ote}: a meal served to all guests at a stated hour and fixed price\\
\verb+ +\hfill --- Merriam-Webster Dictionary
\normalsize
\end{quote}


\section{Introduction}
\label{sec:intro}

Activation functions (AFs) are crucial to the success (or failure) of a neural network (witness the rise in recent years of the ReLU function at the expense of the sigmoid function---see Section~\ref{sec:prev}). With the growing success of deep neural networks over the past decade, AFs have been receiving increased attention, with researchers seeking to design better ones and create networks with better AF layers.

Herein, we propose to take a different route: We then seek ways by which to build AF layers that can outperform a standard neural-network architecture. We call our approach \textit{\`a la carte}, for \textit{\textbf{a}ctivation \textbf{la}yers: \textbf{c}hoose, \textbf{ar}range, \textbf{t}rain, \textbf{e}t voil\`a}.

We propose three methods by which AFs can be combined: 1) randomly, 2) through a state-of-the-art hyper-parameter optimizer with a Tree-structured Parzen Estimator (TPE) sampler, and 3) via said optimizer with a Covariance Matrix Adaptation Evolution Strategy (CMA-ES) sampler.

In the next section we survey the literature on AF design, followed by a presentation of our experimental setup in Section~\ref{sec:setup}.
In Section~\ref{sec:results} we delineate the results and discuss them. Finally, we offer concluding remarks in Section~\ref{sec:conc}.

\section{Activation Functions: Previous Work}
\label{sec:prev}

A recent survey compared trends in AFs both in research and practice of deep learning \cite{nwankpa2018activation}. They examined the following 21 AFs: Sigmoid, HardSigmoid, SiLU, dSiLU, Tanh, Hardtanh, Softmax, Softplus, Softsign, ReLU, LReLU, PReLU, RReLU, SReLU, ELU, PELU, SELU, Maxout, Swish, ELiSH, and HardELiSH (for precise definitions the reader is referred to \cite{nwankpa2018activation}). Highly popular architectures---including AlexNet, ZFNet, VGGNet, SegNet, GoogleNet, SqueezeNet, ResNet, ResNeXt, MobileNets, and SeNet---were shown to have a common AF setup, with Rectified Linear Units (ReLUs) in the hidden layers and Softmax units in the output layer. 
Indeed, in the next section we will set this as our standard architecture for comparison purposes.
\citet{nwankpa2018activation} stated that, ``The most notable observation on the use of AFs for DL applications is that the newer activation functions seem to outperform the older AFs like the ReLU, yet even the latest DL architectures rely on the ReLU function.'' \cite{nwankpa2018activation}

AF approaches can be divided into three categories: tunable or learnable AFs, ensemble methods, and theoretically motivated AFs.

\paragraph{Tunable AFs.}
A very recent survey focused on modern trainable AFs \cite{apicella2021survey}. They proposed a two-class taxonomy comprising fixed-shape AFs and trainable AFs. The former included all the classical AFs, such as sigmoid, tanh, ReLU, and so forth, while the latter class contained AFs whose shape is learned during the training phase. The fixed-shape class was further divided into classic and rectified-based, and the trainable class was further divided into parameterized standard AFs and AFs based on ensemble methods. Finally, they looked at trainable non-standard neuron definitions, which ``change the standard definition of neuron computation but are considered as neural network models with trainable activation functions in the literature'' (e.g., maxout). They noted that in many cases improved performance could be gained by using trainable AFs rather than fixed-shape AFs.

\citet{tavakoli2021splash} introduced SPLASH units (Simple Piecewise Linear and Adaptive with Symmetric Hinges), a class of learnable AFs. SPLASH units are continuous, grounded, and use symmetric hinges that are placed at fixed locations, derived from the data. Compared to nine other learned and fixed AFs, SPLASH units showed superior performance across three datasets and four architectures.

\citet{bohra2020learning} presented an efficient computational solution to train deep neural networks with learnable AFs, specifically focusing on deep spline networks. 

\citet{liu2020reactnet} observed that small variations to activation distributions could greatly affect the semantic feature representations in 1-bit CNNs, and introduced generalized AFs with learnable coefficients. Their proposed network attained good performance at substantially lower computational cost.

\citet{agostinelli2015learning} designed a novel form of piecewise linear AF that was learned independently for each neuron using gradient descent, and showed improvement over deep neural networks composed of static rectified linear units.

\citet{li2013extreme} described an extreme learning machine (ELM) with a tunable AF that was tuned by means of differential evolution based on the input data. 
The tunable AF was of the form $G(S,\alpha)$, with $S$ being the inner product of input and weights combined with bias, and $\alpha$ being the tunable parameter.
They demonstrated improved performance over previous versions of ELM.

\citet{ertuugrul2018novel} presented trained AFs, where an AF was trained for each particular neuron by linear regression. A different AF was generated for each neuron in the hidden layer. The approach was used in random weight artificial neural networks (RWNs), attaining success rates surpassing RWNs that used traditional AFs. 

\citet{scardapane2019kafnets} introduced a novel family of flexible AFs based on an inexpensive kernel expansion at every neuron. They showed that their kernel activation function (KAF) networks outperformed competing approaches on a number of benchmark problems.

Some older papers include \cite{shen2004new}, who introduced a multi-output neural model with a tunable, AF, $f(S,a)$, with $S$ being the  inner product of input and weights, and $a$ being a tunable parameter that could change during training. \citet{piazza1993neural} presented a Look-Up-Table (LUT) AF, offering a preliminary study of an adaptive LUT-based neuron (L-neuron) over some (now-considered) simple problems (XOR, 4-bit parity, and 8-bit parity).

\paragraph{Ensemble methods.}
\citet{maguolo2021ensemble} recently presented an ensemble of convolutional neural networks (CNNs) trained with different AFs, and also introduced a new AF---the Mexican Linear Unit. They showed that an ensemble of multiple CNNs that only differed in the AFs outperformed the results of the single CNNs and of naive ensembles made of ReLU networks.

\cite{nandi2020improving} proposed AF ensembling by majority voting,
wherein a single neural network is trained with five different AFs.
They showed that model accuracy could be improved for four datasets---MNIST, 
Fashion MNIST, Semeion, and ARDIS IV---over methods including CNN, RNN (Recurrent Neural Network), and SVM (Support
Vector Machine).

\paragraph{Theoretically motivated AFs.}
\citet{koccak2021new} introduced a number of new AFs, including: generalized swish, mean-swish, ReLU-swish, triple-state swish, sigmoid-algebraic, triple-state sigmoid, exponential swish, sinc-sigmoid, and derivative of sigmoid AFs. They compared the proposed AFs with some well-known and recently proposed AFs. They found that, ``New AFs can combine the advantages of predefined ones and so they perform better than the older ones.''

\citet{rozsa2019improved} introduced the tent AF and showed that tents make deep learning
models more robust to adversarial attacks. The constraining tent AF has the form:
$f(x;\delta)=\max(0,\delta-\lvert x \rvert)$; as noted by the authors,
``it is basically built from two ReLUs where a ReLU and a horizontally flipped ReLU share the same distance ($\delta$) from the origin.''

\citet{ohn2019smooth} investigated the approximation ability of deep neural networks with a broad class of frequently used AFs. They derived the required depth, width, and sparsity of a deep neural network to approximate any H{\"o}lder smooth function up to a given approximation error.

\citet{misra2019mish} proposed a novel AF called Mish, defined as $f(x)=x \cdot \mathit{tanh}(\mathit{softplus}(x))$, and showed its merits over other AFs on a number of challenging datasets.

\citet{gulcehre2016noisy} proposed a novel method to train neural networks with AFs that strongly saturate when their input
is large. This was achieved by injecting noise to the AF in its saturated regime and learning the level of noise. They showed that their noisy AFs were easier to optimize, and also attained better test errors, since the noise injected to the activations also regularized the model.

\citet{gomes2011comparison} proposed three new AFs: complementary log-log, probit, and log-log. They compared networks using these AFs to networks using classical AFs, over financial time series datasets. They concluded with recommendations of when to apply these new AFs given the type of time series data.

\paragraph{}Having examined the literature we conclude that research into activation functions, much of it very recent, clearly shows a thriving interest in veering away from traditional, popular AFs, attempting to find better networks through improved AFs and AF layers. This line of research is mostly concerned with proposing new forms of AFs, with interesting properties that might improve network performance. 

Rather than designing a new AF we take another direction, proposing to combine a slew of known AFs, and showing how this can be done advantageously. 

\section{Experimental Setup}
\label{sec:setup}

We used Optuna, a state-of-the-art automatic hyper-parameter optimization software framework \cite{akiba2019optuna} (we also considered Hyperopt \cite{bergstra2013hyperopt}, but it is more dated and less potent). Optuna offers a define-by-run style user API where one can dynamically construct the search space, and an efficient sampling algorithm and pruning algorithm. Moreover, our experience has shown it to be fairly easy to set up. 
Optuna formulates the hyper-parameter optimization problem as a process of minimizing or maximizing an objective function given a set of hyper-parameters as an input. 

As noted by \citet{akiba2019optuna}, there are generally two types of sampling methods: relational sampling, which exploits correlations amongst parameters, and independent sampling, which samples each parameter independently. 
Optuna offers two main samplers, both of which we used herein.
Covariance Matrix Adaptation Evolution Strategy (CMA-ES) performs relational sampling, while
Tree-structured Parzen Estimator (TPE) performs independent sampling. 
Optuna also provides pruning: automatic early stopping of unpromising trials \cite{akiba2019optuna}.

We implemented our experiments in PyTorch, a popular deep learning framework \cite{paszke2019pytorch}. We selected 44 potential AFs, some of which are ``officially'' defined in \texttt{torch.nn} as AFs (e.g., ReLU and Sigmoid), while the others are simply mathematical functions over tensors (e.g., Abs and Sin). Note that where the latter are concerned there might be some question as to their differentiability (PyTorch performs automatic differentiation---indeed this feature is one of its main strengths). However, in practice, this does not pose a problem since the number of points at which the function is non-differentiable is usually (infinitely) small. For example, ReLU is non-differentiable at zero, however, in practice, it is rare to stumble often onto this value, and when we do, an arbitrary value can be assigned (e.g., zero), with little to no effect on performance.

We also added four top-performing AFs from the literature discussed in Section~\ref{sec:prev}. This served the twofold purpose of testing newly proposed AFs and demonstrating the ease with which our framework can accommodate new AF ideas. The full list of AFs is given in Figure~\ref{fig:AFs}.

\begin{figure}
\begin{mdframed}[style=mdstyle]
\small
\textbf{PyTorch AFs}: \texttt{ELU}, \texttt{Hardshrink}, \texttt{Hardtanh}, \texttt{LeakyReLU}, \texttt{LogSigmoid}, \texttt{PReLU}, \texttt{ReLU}, \texttt{ReLU6}, \texttt{RReLU}, \texttt{SELU}, \texttt{CELU}, \texttt{GELU}, \texttt{Sigmoid}, \texttt{Softplus}, \texttt{Softshrink}, \texttt{Softsign}, \texttt{Tanh}, \texttt{Tanhshrink}, \texttt{Softmin}, \texttt{Softmax}, \texttt{LogSoftmax}\\

\textbf{PyTorch Math}: \texttt{Abs}, \texttt{Acos}, \texttt{Angle}, \texttt{Asin}, \texttt{Atan}, \texttt{Ceil}, \texttt{Cos}, \texttt{Cosh}, \texttt{Digamma}, \texttt{Erf}, \texttt{Erfc}, \texttt{Exp}, \texttt{Floor}, \texttt{Frac}, \texttt{GumbelSoftmax}, \texttt{Log}, \texttt{Log10}, \texttt{Neg}, \texttt{Round}, \texttt{Sin}, \texttt{Sinh}, \texttt{Tan}, \texttt{Trunc}\\

\textbf{New AFs culled from the literature}:

\texttt{Mish}: $f(x)=x \cdot \mathit{tanh}(\mathit{softplus}(x))$ \cite{misra2019mish}

\texttt{Generalized Swish}: $f(x)=x \cdot \mathit{sigmoid}(e^{-x})$ \cite{koccak2021new}

\texttt{Sigmoid Derivative}: $f(x)=e^{-x} \cdot (\mathit{sigmoid}(x))^2$ \cite{koccak2021new}

\texttt{CLogLogM}: $f(x)=1 - 2 \cdot e^{-0.7 \cdot e^x}$   \cite{gomes2011comparison}
\end{mdframed}
\normalsize
\caption{List of 48 AFs used herein. For full descriptions the reader is referred to \protect\url{pytorch.org} and the respective cited papers.}
\label{fig:AFs}
\end{figure}

For our experiments we selected classification datasets from \url{OpenML.org} \cite{OpenML2013}, which boasts over 21,000 datasets (and is constantly growing). Upon examination we found 465 classification datasets with at least 10,000 samples. Of these we selected 25 datasets, such that we ended up with a variety of number of samples, features, and classes. 

For each dataset we performed 30 replicate runs with 5-layer networks and 30 replicate runs with 10-layer networks. While Optuna could have been tasked with optimizing the number of layers as well, we chose not to increase the already-large search space, and focus on AFs alone---our main interest in this paper. We thus opted for 5- and 10-layer networks, values chosen so as to eschew shallow networks while still affording extensive testing of our approach through multiple replicates over multiple datasets. 

Each network computational layer was fully connected, and performed the standard linear computation, $y=Wx+b$, with $W$ being the weight matrix, $x$ being the input vector, and $b$ being the bias (\texttt{torch.nn.Linear}). This was followed by the AF (layer) computation.
We used the Adam optimizer \cite{kingma2014adam} with a learning rate of $0.001$, cross entropy loss, and a single batch (i.e., the entire training set).
The hidden-layer sizes were set to 64 nodes, with the preceding input and succeeding output dimensions depending on the particular dataset.

The pseudo-code of the experimental setup is given in Algorithm~\ref{alg:setup}. 
Each replicate run began with a 70\%-30\% training/test split of the dataset.
We fit scikit-learn's \cite{scikit-learn} \texttt{StandardScaler} to the training set and applied the fitted scaler to the test set. This ensured that features had zero mean and unit variance (often helpful where neural networks are concerned).

\begin{algorithm*}
\footnotesize
\caption{Experimental setup (per dataset)}\label{alg:setup}
\begin{algorithmic}[1]
\Statex
\Require
\Indent
\Statex \textit{dataset} $\gets$ dataset to be used
\EndIndent

\Ensure
\Indent
\Statex Test scores (for each network type, over all replicates)
\EndIndent

\Statex
\For{\textit{rep} $\gets$ 1 to \textit{30}} 
    \State Randomly split \textit{dataset} into 70\% \textit{training set} and 30\% \textit{test set}
    \State Fit \texttt{StandardScaler} to \textit{training set} and apply fitted scaler to \textit{test set}
    \State Train a standard network over \textit{training set} and test over \textit{test set}
    \State Independently generate 1000 networks with random AFs, train each one over \textit{training set}, test each one over \textit{test set}, retain top test score
    \State Run Optuna with TPE sampler for 1000 trials over \textit{training set}, test returned best model over \textit{test set}
    \State Run Optuna with CMA-ES sampler for 1000 trials over \textit{training set}, test returned best model over \textit{test set}
\EndFor
\end{algorithmic}
\normalsize
\end{algorithm*} 

We then executed and compared four different AF architectures:
\begin{enumerate}
    \item The standard architecture (as seen in Section~\ref{sec:prev}): hidden ReLU units with a Softmax output. Train the network over the training set and record the score over the test set as the replicate's score. The test-set score equalled prediction accuracy, with the best value being 1 and the worst value being 0. 
    \item Generate 1000 random networks. A random network is generated by selecting at random with uniform probability 5 or 10 AFs (for 5 or 10 layers, respectively) from the list of 48 AFs (Figure~\ref{fig:AFs}). Train each random network over the training set and test it over the test set. Record the best test score of the 1000 random networks as the replicate's score.
    \item Run Optuna for 1000 trials with a TPE sampler. Set Optuna to optimize as hyper-parameters the (5 or 10) AFs (for 5 or 10 AF layers). A single trial comprises training the network with a particular set of hyper-paremeters supplied by Optuna (a list of AFs) over the training set. Test the best network returned by Optuna (after 1000 trials) over the test set, and record this score as the replicate's score.
    \item As in the previous item, except use Optuna with the CMA-ES sampler.
\end{enumerate}

A network was trained for up to 300 epochs, with early stopping: exit if accuracy improvement from one epoch to the next is less than 0.001 for 10 successive epochs. 
The code is available at \url{github.com/moshesipper}.

\citet{basirat2018quest} came closest to what we propose above. They suggested the use of a genetic algorithm to  combine 11 AFs through the use of 8 operators: addition, subtraction, multiplication, division, exponentiation, minimum, maximum, and function composition. 
They showed that simple combinations of AFs could be evolved.
Their approach differed in the following from ours: 1) A network was tested with a single AF, whereas our method is based on a combination of AFs; 2) they tested their approach on a smaller number of datasets (3 vs. our 25); (3) our set of potential AFs is significantly larger, with far more mathematical operators (which turn out to be useful, as we shall see below); (4) our AF-layout design employs both a less powerful method (random)  and a more sophisticated method (Optuna) than the genetic algorithm used therein.

\section{Results and Discussion}
\label{sec:results}

Table~\ref{tab:results} shows the results of all 50 experiments, each row summarizing 30 replicate runs. 
We performed two tests to asses statistical significance:
1) a 10,000-round permutation test comparing the medians of the respective method's scores (random AFs or Optuna-designed) with the standard network's scores; and 
2) a 10,000-round permutation test comparing the medians of the top method's scores with the second-best network's scores.

\begin{table}
\centering
\caption{\footnotesize Experimental results. All scores shown are those of the test sets.
\textit{dataset}: OpenML dataset id;
\textit{samp}: number of samples; 
\textit{feat}: number of features;
\textit{cls}: number of classes;
\textit{lay}: number of network layers;
\textit{standard}: median score over 30 replicates of standard network;
\textit{random}: median score over 30 replicates of best random network score (of 1000) per replicate;
\textit{tpe}: median score over 30 replicates of best network per replicate found by Optuna using TPE sampler;
\textit{cmaes}: median score over 30 replicates of best network per replicate found by Optuna using CMA-ES sampler;
\textit{top}: method that produced the top median score over 30 replicates. 
For \textit{random}, \textit{tpe}, \textit{cmaes}: we performed a 10,000-round permutation test comparing the medians of the respective method's scores with the standard network's scores. 
For \textit{top}: we performed a 10,000-round permutation test comparing the medians of the top method's scores with the second-best network's scores.
A `!!' denotes a p-value $<0.001$ and a `!' denotes a p-value $<0.05$ (otherwise p-value $>=0.05$).
}
\label{tab:results}

\ssmall
\begin{tabular}{r|c|c|c|c|c|c|c|c|c}
dataset              & samp                 & feat               & cls               & lay & standard & random   & tpe      & cmaes    & top \\ \hline
\multirow{2}*{6} & \multirow{2}*{20000} & \multirow{2}*{16}  & \multirow{2}*{26} & 5 & 0.119 & 0.924 !! & 0.931 !! & 0.923 !! & tpe !  \\
& & & & 10 & 0.04 & 0.829 !! & 0.896 !! & 0.817 !! & tpe !!  \\
\multirow{2}*{32} & \multirow{2}*{10992} & \multirow{2}*{16}  & \multirow{2}*{10} & 5 & 0.436 & 0.986 !! & 0.986 !! & 0.984 !! & random  \\
& & & & 10 & 0.103 & 0.979 !! & 0.982 !! & 0.976 !! & tpe !  \\
\multirow{2}*{279} & \multirow{2}*{45164} & \multirow{2}*{74}  & \multirow{2}*{11} & 5 & 0.509 & 0.961 !! & 0.966 !! & 0.96 !! & tpe !!  \\
& & & & 10 & 0.509 & 0.566 !! & 0.623 !! & 0.574 !! & tpe  \\
\multirow{2}*{1044} & \multirow{2}*{10936} & \multirow{2}*{27}  & \multirow{2}*{3} & 5 & 0.547 & 0.591 !! & 0.53 & 0.483 & random !!  \\
& & & & 10 & 0.387 & 0.513 !! & 0.374 ! & 0.422 !! & random !!  \\
\multirow{2}*{1459} & \multirow{2}*{10218} & \multirow{2}*{7}  & \multirow{2}*{10} & 5 & 0.206 & 0.687 !! & 0.737 !! & 0.675 !! & tpe !!  \\
& & & & 10 & 0.121 & 0.59 !! & 0.625 !! & 0.577 !! & tpe !  \\
\multirow{2}*{1471} & \multirow{2}*{14980} & \multirow{2}*{14}  & \multirow{2}*{2} & 5 & 0.551 & 0.878 !! & 0.859 !! & 0.852 !! & random  \\
& & & & 10 & 0.55 & 0.721 !! & 0.727 !! & 0.723 !! & tpe  \\
\multirow{2}*{1476} & \multirow{2}*{13910} & \multirow{2}*{128}  & \multirow{2}*{6} & 5 & 0.964 & 0.991 !! & 0.991 !! & 0.989 !! & tpe  \\
& & & & 10 & 0.21 & 0.983 !! & 0.988 !! & 0.983 !! & tpe !!  \\
\multirow{2}*{1477} & \multirow{2}*{13910} & \multirow{2}*{129}  & \multirow{2}*{6} & 5 & 0.975 & 0.991 !! & 0.992 !! & 0.99 !! & tpe  \\
& & & & 10 & 0.212 & 0.982 !! & 0.988 !! & 0.983 !! & tpe !!  \\
\multirow{2}*{1478} & \multirow{2}*{10299} & \multirow{2}*{561}  & \multirow{2}*{6} & 5 & 0.964 & 0.98 !! & 0.974 !! & 0.976 !! & random !!  \\
& & & & 10 & 0.192 & 0.971 !! & 0.971 !! & 0.969 !! & random  \\
\multirow{2}*{1481} & \multirow{2}*{28056} & \multirow{2}*{6}  & \multirow{2}*{18} & 5 & 0.149 & 0.637 !! & 0.699 !! & 0.642 !! & tpe !!  \\
& & & & 10 & 0.127 & 0.463 !! & 0.549 !! & 0.41 !! & tpe !!  \\
\multirow{2}*{1531} & \multirow{2}*{10176} & \multirow{2}*{3}  & \multirow{2}*{5} & 5 & 0.962 & 0.966 !! & 0.966 !! & 0.965 !! & tpe  \\
& & & & 10 & 0.961 & 0.962 & 0.962 & 0.962 & tpe  \\
\multirow{2}*{1532} & \multirow{2}*{10668} & \multirow{2}*{3}  & \multirow{2}*{5} & 5 & 0.964 & 0.968 !! & 0.967 ! & 0.967 !! & random  \\
& & & & 10 & 0.964 & 0.964 & 0.964 & 0.964 & cmaes  \\
\multirow{2}*{1533} & \multirow{2}*{10386} & \multirow{2}*{3}  & \multirow{2}*{5} & 5 & 0.964 & 0.968 !! & 0.967 !! & 0.967 ! & random  \\
& & & & 10 & 0.963 & 0.964 & 0.964 & 0.964 & cmaes  \\
\multirow{2}*{1534} & \multirow{2}*{10190} & \multirow{2}*{3}  & \multirow{2}*{5} & 5 & 0.963 & 0.967 !! & 0.966 ! & 0.965 ! & random  \\
& & & & 10 & 0.962 & 0.962 & 0.962 & 0.962 & tpe  \\
\multirow{2}*{1536} & \multirow{2}*{10130} & \multirow{2}*{3}  & \multirow{2}*{5} & 5 & 0.962 & 0.967 !! & 0.966 !! & 0.966 !! & random  \\
& & & & 10 & 0.962 & 0.963 & 0.963 & 0.962 & tpe  \\
\multirow{2}*{1537} & \multirow{2}*{28626} & \multirow{2}*{3}  & \multirow{2}*{5} & 5 & 0.974 & 0.976 ! & 0.975 ! & 0.975 & random  \\
& & & & 10 & 0.975 & 0.975 & 0.975 & 0.975 & tpe  \\
\multirow{2}*{1568} & \multirow{2}*{12958} & \multirow{2}*{8}  & \multirow{2}*{4} & 5 & 0.649 & 0.985 !! & 0.992 !! & 0.984 !! & tpe !!  \\
& & & & 10 & 0.329 & 0.979 !! & 0.992 !! & 0.978 !! & tpe !!  \\
\multirow{2}*{4154} & \multirow{2}*{14240} & \multirow{2}*{30}  & \multirow{2}*{2} & 5 & 0.998 & 1.0 !! & 0.999 !! & 0.999 !! & random  \\
& & & & 10 & 0.998 & 0.998 & 0.998 & 0.998 & standard  \\
\multirow{2}*{4534} & \multirow{2}*{11055} & \multirow{2}*{30}  & \multirow{2}*{2} & 5 & 0.938 & 0.958 !! & 0.96 !! & 0.956 !! & tpe  \\
& & & & 10 & 0.555 & 0.938 !! & 0.946 !! & 0.936 !! & tpe !  \\
\multirow{2}*{40985} & \multirow{2}*{45781} & \multirow{2}*{2}  & \multirow{2}*{20} & 5 & 0.062 & 0.068 !! & 0.061 ! & 0.062 & random !!  \\
& & & & 10 & 0.063 & 0.067 !! & 0.062 ! & 0.062 & random !!  \\
\multirow{2}*{41027} & \multirow{2}*{44819} & \multirow{2}*{6}  & \multirow{2}*{3} & 5 & 0.52 & 0.825 !! & 0.855 !! & 0.824 !! & tpe !!  \\
& & & & 10 & 0.514 & 0.776 !! & 0.795 !! & 0.773 !! & tpe !  \\
\multirow{2}*{41526} & \multirow{2}*{15547} & \multirow{2}*{60}  & \multirow{2}*{2} & 5 & 0.489 & 0.511 !! & 0.51 !! & 0.511 !! & random  \\
& & & & 10 & 0.486 & 0.514 !! & 0.513 !! & 0.513 !! & random  \\
\multirow{2}*{41671} & \multirow{2}*{20000} & \multirow{2}*{20}  & \multirow{2}*{5} & 5 & 0.559 & 0.569 !! & 0.581 !! & 0.572 !! & tpe  \\
& & & & 10 & 0.559 & 0.56 & 0.56 & 0.56 & random  \\
\multirow{2}*{42641} & \multirow{2}*{18982} & \multirow{2}*{79}  & \multirow{2}*{5} & 5 & 0.858 & 0.952 !! & 0.955 !! & 0.95 !! & tpe !  \\
& & & & 10 & 0.278 & 0.904 !! & 0.928 !! & 0.89 !! & tpe  \\
\multirow{2}*{42670} & \multirow{2}*{18982} & \multirow{2}*{79}  & \multirow{2}*{5} & 5 & 0.855 & 0.952 !! & 0.953 !! & 0.951 !! & tpe  \\
& & & & 10 & 0.276 & 0.894 !! & 0.941 !! & 0.902 !! & tpe !  \\
\end{tabular}
\normalsize

\end{table}

At least one of our newly proposed methods (random network, Optuna with TPE sampler, Optuna with CMA-ES sampler) attained statistically significant improvement in 42 of the 50 experiments. In most cases \textit{all} three methods were better. 
Moreover, in several cases improvement was not merely statistically significant, but indeed twofold, threefold, and even more.

A statistically significant number-one ranking (when compared with the second-best) was attained by Optuna using a TPE sampler 17 times, and by the random network 5 times.
AF layout design with Optuna would seem to be an excellent choice when one wishes to improve a deep learner's performance.

In addition to the results given in Table~\ref{tab:results}, we were now in possession of 4500 architectures (25 datasets $\times$ 30 replicates $\times$ 2 layer counts $\times$ 3 non-standard methods). Given the large search space of architectures ($48^5$ for 5 layers and $48^{10}$ for 10 layers) it was not surprising that almost all 4500 architectures were unique (the only exceptions were 2 5-layer networks that appeared twice). 
We could, however, take a closer at the 33750 AFs within these architectures, and examine their frequencies---which are given in Table~\ref{tab:af-freqs}.

\begin{table}
\caption{\footnotesize Frequencies of top-10 AFs in the best networks found by the three methods discussed in the text (random, Optuna with TPE sampler, Optuna with CMA-ES sampler).
Shown by category: AFs in input layer, AFs in hidden layers, and AFs in output layer.}
\label{tab:af-freqs}
\centering

\scriptsize
\begin{tabular}{rl|rl|rl}
\multicolumn{6}{c}{\textbf{5 layers}}  \\ 
\multicolumn{2}{c|}{input} & \multicolumn{2}{c|}{hidden} & \multicolumn{2}{c}{output} \\  
AF & Freq & AF & Freq & AF & Freq  \\ \hline
Softshrink & 6.7\%  &  Softshrink & 6.2\%  &  Exp & 11.1\%  \\
Sinh & 6.3\%  &  Abs & 5.0\%  &  Sinh & 8.1\%  \\
Abs & 5.8\%  &  Sinh & 4.5\%  &  Cosh & 7.5\%  \\
SELU & 4.9\%  &  Hardshrink & 4.5\%  &  LogSoftmax & 4.4\%  \\
Angle & 4.5\%  &  Hardtanh & 4.5\%  &  Neg & 4.0\%  \\
Neg & 4.4\%  &  Exp & 4.1\%  &  Abs & 4.0\%  \\
Tanhshrink & 3.7\%  &  Sin & 4.0\%  &  Softplus & 3.7\%  \\
Hardshrink & 3.5\%  &  Erf & 3.8\%  &  PReLU & 3.6\%  \\
Exp & 3.5\%  &  SELU & 3.6\%  &  SELU & 3.6\%  \\
Hardtanh & 3.4\%  &  ReLU6 & 3.2\%  &  RReLU & 3.5\%  \\
\end{tabular}

\vspace{10pt}

\begin{tabular}{rl|rl|rl}
\multicolumn{6}{c}{\textbf{10 layers}} \\
\multicolumn{2}{c|}{input} & \multicolumn{2}{c|}{hidden} & \multicolumn{2}{c}{output} \\
AF & Freq & AF & Freq & AF & Freq  \\ \hline
Neg & 3.8\%  &  SELU & 3.7\%  &  GumbelSoftmax & 9.1\%  \\
Sinh & 3.8\%  &  Sin & 3.5\%  &  Exp & 5.4\%  \\
Exp & 3.7\%  &  Neg & 3.2\%  &  LogSoftmax & 4.2\%  \\
Angle & 3.5\%  &  Erf & 3.2\%  &  CELU & 4.0\%  \\
Sin & 3.4\%  &  Hardtanh & 3.1\%  &  ELU & 4.0\%  \\
SELU & 3.3\%  &  Exp & 3.0\%  &  PReLU & 3.8\%  \\
Hardtanh & 3.3\%  &  CELU & 3.0\%  &  Cosh & 3.7\%  \\
Erf & 3.2\%  &  Tanh & 3.0\%  &  SELU & 3.7\%  \\
Tanh & 3.1\%  &  Atan & 3.0\%  &  Neg & 3.5\%  \\
GeneralizedSwish & 2.9\%  &  Sinh & 3.0\%  &  RReLU & 3.3\%  \\
\end{tabular}
\normalsize

\end{table}

Examining Table~\ref{tab:af-freqs}, we can make several interesting observations. Firstly, we note what is \textit{not} there: the commonly used ReLU and Softmax functions (they do appear lower down the list, not shown for brevity). We note the preponderance of usually overlooked mathematical functions, which would seem to be a good choice for AF candidates, e.g., sine, hyperbolic sine, negative, and exponential. The diversity of AFs chosen by the methods presented herein is also apparent. Providing a large set of AF candidates as grist for the AF layout-design mill is beneficial and can be fruitfully applied. Of the four newly proposed AFs (Figure~\ref{fig:AFs}) only \texttt{Generalized Swish} made it into one of the top-10 rankings. 

We asked whether for particular datasets specific AFs appeared more frequently, an inquiry that led to Table~\ref{tab:dsfreqs}.
We note that for some datasets certain AFs were selected with a notably higher frequency.
And once again we note the preponderance of AFs not often used.
 
\begin{table}
\caption{\footnotesize Frequencies of topmost AF in the best networks found by all three methods, by dataset.
\textit{ds}: OpenML dataset id;
\textit{lay}: number of network layers.
Shown by category: AFs in input layer, AFs in hidden layers, and AFs in output layer.}
\label{tab:dsfreqs}
\centering

\ssmall
\begin{tabular}{cc|rl|rl|rl}
ds & lay & AF & Freq & AF & Freq & AF & Freq  \\ \hline
& & \multicolumn{2}{c|}{input} & \multicolumn{2}{c|}{hidden} & \multicolumn{2}{c}{output} \\  \cline{3-8}
\multirow{2}*{6} & 5 & Sinh & 13.3\% & Exp & 11.9\% & Exp & 24.4\% \\
 & 10 & Tanh & 10.0\% & SELU & 5.7\% & CELU & 12.2\% \\ \thinhline
\multirow{2}*{32} & 5 & Exp & 18.9\% & Exp & 12.2\% & Sinh & 14.4\% \\
 & 10 & Sinh & 10.0\% & Exp & 5.3\% & CELU & 11.1\% \\ \thinhline
\multirow{2}*{279} & 5 & Sinh & 7.8\% & Sinh & 8.1\% & Exp & 25.6\% \\
 & 10 & Sin & 5.6\% & Digamma & 5.3\% & SigmoidDerivative & 11.1\% \\ \thinhline
\multirow{2}*{1044} & 5 & Round & 17.8\% & Abs & 15.2\% & CELU & 11.1\% \\
 & 10 & ReLU & 6.7\% & SELU & 4.4\% & GumbelSoftmax & 28.9\% \\ \thinhline
\multirow{2}*{1459} & 5 & Angle & 43.3\% & Exp & 9.6\% & Softplus & 20.0\% \\
 & 10 & Angle & 7.8\% & Angle & 6.8\% & CELU & 8.9\% \\ \thinhline
\multirow{2}*{1471} & 5 & Angle & 13.3\% & Abs & 11.9\% & Atan & 11.1\% \\
 & 10 & ReLU6 & 5.6\% & Abs & 5.1\% & GumbelSoftmax & 40.0\% \\ \thinhline
\multirow{2}*{1476} & 5 & Abs & 28.9\% & Erfc & 10.0\% & Exp & 16.7\% \\
 & 10 & LeakyReLU & 11.1\% & SELU & 6.2\% & Exp & 11.1\% \\ \thinhline
\multirow{2}*{1477} & 5 & Abs & 31.1\% & Abs & 9.6\% & Exp & 18.9\% \\
 & 10 & Erf & 7.8\% & ELU & 5.1\% & ELU & 11.1\% \\ \thinhline
\multirow{2}*{1478} & 5 & Cos & 30.0\% & Cos & 6.3\% & Exp & 15.6\% \\
 & 10 & Cos & 12.2\% & Abs & 6.4\% & GumbelSoftmax & 11.1\% \\ \thinhline
\multirow{2}*{1481} & 5 & Sinh & 23.3\% & Sin & 10.7\% & Exp & 16.7\% \\
 & 10 & CELU & 7.8\% & Exp & 4.6\% & Cosh & 15.6\% \\ \thinhline
\multirow{2}*{1531} & 5 & Softshrink & 13.3\% & Softshrink & 11.9\% & SELU & 8.9\% \\
 & 10 & Trunc & 6.7\% & Digamma & 3.9\% & CELU & 5.6\% \\ \thinhline
\multirow{2}*{1532} & 5 & Tanhshrink & 12.2\% & Softshrink & 11.5\% & Neg & 8.9\% \\
 & 10 & Sinh & 8.9\% & Digamma & 5.1\% & Mish & 7.8\% \\ \thinhline
\multirow{2}*{1533} & 5 & Tanhshrink & 14.4\% & Hardshrink & 10.0\% & Abs & 6.7\% \\
 & 10 & Exp & 5.6\% & Ceil & 3.5\% & ELU & 6.7\% \\ \thinhline
\multirow{2}*{1534} & 5 & Tanhshrink & 11.1\% & Softshrink & 12.6\% & PReLU & 10.0\% \\
 & 10 & Cosh & 7.8\% & Digamma & 3.8\% & Softshrink & 6.7\% \\ \thinhline
\multirow{2}*{1536} & 5 & Tanhshrink & 12.2\% & Hardshrink & 10.7\% & LogSoftmax & 10.0\% \\
 & 10 & Cosh & 6.7\% & Digamma & 5.1\% & Softshrink & 6.7\% \\ \thinhline
\multirow{2}*{1537} & 5 & Sinh & 13.3\% & Softshrink & 9.6\% & ReLU & 8.9\% \\
 & 10 & Exp & 8.9\% & Digamma & 3.6\% & ReLU & 6.7\% \\ \thinhline
\multirow{2}*{1568} & 5 & Hardtanh & 8.9\% & Abs & 13.3\% & Sinh & 27.8\% \\
 & 10 & Exp & 8.9\% & Atan & 4.7\% & Exp & 11.1\% \\ \thinhline
\multirow{2}*{4154} & 5 & Softshrink & 16.7\% & Hardshrink & 12.2\% & GumbelSoftmax & 20.0\% \\
 & 10 & Abs & 11.1\% & Neg & 3.8\% & Floor & 5.6\% \\ \thinhline
\multirow{2}*{4534} & 5 & Abs & 40.0\% & Abs & 11.9\% & Abs & 12.2\% \\
 & 10 & RReLU & 5.6\% & Sin & 6.0\% & GumbelSoftmax & 35.6\% \\ \thinhline
\multirow{2}*{40985} & 5 & Tan & 22.2\% & Hardtanh & 8.1\% & GumbelSoftmax & 23.3\% \\
 & 10 & Digamma & 11.1\% & Digamma & 5.6\% & GumbelSoftmax & 18.9\% \\ \thinhline
\multirow{2}*{41027} & 5 & Exp & 21.1\% & Softshrink & 8.9\% & Neg & 11.1\% \\
 & 10 & Hardtanh & 8.9\% & Erf & 4.7\% & GumbelSoftmax & 40.0\% \\ \thinhline
\multirow{2}*{41526} & 5 & Softshrink & 34.4\% & Softshrink & 22.6\% & Softmin & 5.6\% \\
 & 10 & Softshrink & 13.3\% & Softshrink & 12.2\% & GumbelSoftmax & 11.1\% \\ \thinhline
\multirow{2}*{41671} & 5 & SELU & 12.2\% & SELU & 9.6\% & GumbelSoftmax & 16.7\% \\
 & 10 & Cosh & 6.7\% & GumbelSoftmax & 6.1\% & GumbelSoftmax & 8.9\% \\ \thinhline
\multirow{2}*{42641} & 5 & Softshrink & 10.0\% & Abs & 6.7\% & Exp & 28.9\% \\
 & 10 & Angle & 8.9\% & Erf & 6.1\% & Exp & 12.2\% \\ \thinhline
\multirow{2}*{42670} & 5 & Softshrink & 11.1\% & Abs & 8.9\% & Exp & 22.2\% \\
 & 10 & Angle & 10.0\% & SELU & 7.4\% & PReLU & 11.1\% \\ \thinhline
\end{tabular}

\normalsize
\end{table}

The freedom to choose from a large set of possible AFs and compose AF layouts can produce better-performing networks. It would seem to help reduce error from a model by avoiding overfitting to the training set.
As such, it might be acting as a regularizer, performing a beneficial tradeoff by reducing variance significantly without increasing bias (substantially). This regularization effect of \`a la carte is a conjecture at this point.

\section{Concluding Remarks}
\label{sec:conc}
Presenting deep networks \`a la carte, we showed that it is possible to significantly improve performance by randomly choosing non-standard AF architectures. And even better results can be obtained by using a state-of-the-art hyper-parameter optimizer.

One might inquire as to the cost of searching for better AF layouts. Deep networks come with a plethora of hyper-parameters that need to be tuned, and adding AFs into the mix could therefore be cost-effective when compared with the possible benefits. Our experience has shown that Optuna is efficient at hyper-parameter optimization in general, and at optimizing AF layouts in particular.
Ultimately, whether the cost of \`a la carte is worthwhile or not depends on how important it is for the user to improve performance for a particular task.  

We believe this work may have broad implications as it provides an automated means to attain better deep-learning models, with little input from the user (except, possibly, adding AF candidates to our own plethora).
We can further suggest a number of avenues for future research:
\begin{itemize}
    \item Devise other methods for designing AF architectures.
    \item Seek a better understanding of the dynamics involved. For example, does the approach indeed perform some form of regularization, as suggested in Section~\ref{sec:results}?
    \item While we noted that virtually all AF layouts were unique, and were able to perform an analysis of AF frequencies, it would be of interest to carry this further. No AF is an island: possibly, certain AFs perform better in the presence of certain others.  
    \item There might be additional functions that could be beneficially added to the ``bag of AFs'' (Figure~\ref{fig:AFs}).
    \item Apply \`a la carte to other types of networks, e.g., convolutional neural networks and recurrent neural networks.
    \item Go beyond AF selection---to AF construction. As we saw in Section~\ref{sec:prev}, researchers have suggested tailored combinations of previous AFs (or functions that are not AFs), and \citet{basirat2018quest} showed that evolving simple, new AFs is possible. Perhaps this AF-design process could be further enhanced, as we did for difficult design problems, such as designing game strategies \cite{sipper2007designing} and designing novel statistics \cite{moore2019automated}.
\end{itemize}

\section*{Acknowledgements}
I thank Zvika Haramaty, Raz Lapid, and Snir Vitrack Tamam for helpful comments.
\footnotesize 

\bibliographystyle{spmpsci}
\bibliography{all_refs}
\end{document}